\documentclass[letterpaper]{article} 
\usepackage[hyperref]{emnlp2020}
\usepackage{times}  
\usepackage{graphicx}  
\usepackage{qtree}
\usepackage{natbib}
\usepackage{multirow}

\usepackage{latexsym}

\usepackage{microtype}
\usepackage{subcaption}

\graphicspath{{Figures/}}

\aclfinalcopy

\begin{document}

\title{Converting the Point of View of\\Messages Spoken to Virtual Assistants
}

\author{
\centering 
\begin{tabular}{c c c c c c} 
Isabelle G. Lee  & Vera Zu & Sai Srujana Buddi & Dennis Liang & Purva Kulkarni & Jack G.M. FitzGerald\\
\multicolumn{6}{c}{Amazon Alexa} \\
\multicolumn{6}{c}{ \{ isabelll, xinzu, buddisa, liangxin, kulkpurv, jgmf \}@amazon.com }
\end{tabular}

}

\maketitle

\begin{abstract}
Virtual Assistants can be quite literal at times.
If a user says \textit{tell Bob I love him,} most virtual assistants will extract the message \textit{I love him} and send it to the user's contact named Bob, rather than properly converting the message to \textit{I love you}.
We designed a system that takes a voice message from one user, converts the point of view of the message, and then delivers the result to its target user.
We developed a rule-based model, which integrates a linear text classification model, part-of-speech tagging, and constituency parsing with rule-based transformation methods.
We also investigated Neural Machine Translation (NMT) approaches, including traditional recurrent networks, CopyNet, and T5.
We explored 5 metrics to gauge both naturalness and  faithfulness automatically, and we chose to use BLEU plus METEOR for faithfulness, as well as relative perplexity using a separately trained language model (GPT) for naturalness.
Transformer-Copynet and T5 performed similarly on faithfulness metrics, with T5 scoring 63.8 for BLEU and 83.0 for METEOR.
CopyNet was the most natural, with a relative perplexity of 1.59.
CopyNet also has 37 times fewer parameters than T5.
We have publicly released our dataset, which is composed of 46,565 crowd-sourced samples.

\end{abstract}

\section{Introduction}\label{intro}

Virtual Assistants (VAs), such as Amazon Alexa or Google Assistant, are cloud-based software systems that ingest spoken utterances from individual users, detect the users' intents from the utterances, and perform the desired tasks \citep{kongthon2009implementing, tunstall2014enhanced, memeti2018papa}.
Common VA tasks include playing music, setting timers, answering encyclopedic questions, and controlling smart home devices \citep{lopez2017alexa}.
In addition, VAs are used for communication between two users.
With the help of VAs, users can make and receive calls, as well as send a voice message or text message to their contacts.
This paper focuses on voice messaging.

Messaging can be implemented in such a way that the messages are snipped from the utterance in their spoken form \citep{mohit2014named}.
For example, a user may say \textit{tell Bob that I'm running late}.
The Named Entity Recognition (NER) model could extract \textit{I'm running late} as the message content and pass that to the recipient directly.
Such an approach works in many cases, but it does not perform well if the user says something like \textit{ask bob if he's coming for dinner}, for which the recipient would receive \textit{(if) he's coming for dinner} using a simple NER snipping approach.
In this way, direct messages are distinguished from indirect messages \citep{li1986direct}.
Indirect messages require further natural language processing (NLP) to convert the point of view (POV).
Crucially, the model needs to identify the co-reference relation between Bob, the recipient, and the anaphor \textit{he}.

Additional difficulty arises when we use the VA as an intermediary between the source and the recipient of the message, not simply a voice machine.
Instead of reciting the message \textit{I'm running late} verbatim to Bob, to achieve natural, believable human-robot interaction, the VA should say something like \textit{Joe says he's running late} or simply \textit{Joe is running late}.
That is, for the VA to become a true intermediary in the conversation, the POV conversion must apply to direct messages as well.

The VA-assisted conversation is exemplified in Table \ref{conversationexamples}.
It's a two step process\textemdash messages are relayed from a source to the VA, and then from the VA to a recipient.
The message and its context (e.g., who dictates the message, when and where) are interpreted by the VA, undergo POV conversion, and then are reproduced for the recipient.

\begin{table}[h!]
\caption{Examples of Human-VA Interaction}
\label{conversationexamples}
\begin{tabular}{ll}
\hline
\multicolumn{2}{c}{Example 1} \\
\hline
Joe$\rightarrow$VA& Tell bob I'm running late \\  
VA$\rightarrow$Bob& Joe says he's running late \\ 
\hline
\multicolumn{2}{c}{Example 2} \\
\hline
Joe$\rightarrow$VA& Ask Bob if he's coming for dinner \\  
VA$\rightarrow$Bob& Joe asks if you are coming for dinner  \\ 
\hline
\end{tabular}
\end{table}

%

The most direct approach to solving POV conversion is to author a suite of rules for grammatical conversion. These rules could be used in conjunction with the named entity recognition that is already being performed by the VA. A rule-based approach is deterministic and easily traceable. If operationalized, it would be trivial to debug the model's behavior.

There are a few disadvantages of a rule-based approach, however. First, it requires that someone must hand-author the rules, which is particularly burdensome as the number of languages scale. Second, depending on the types of rules implemented, a rule-based approach may output converted utterances that sound robotic or unnatural. For these reasons, we also considered encoder-decoder approaches, which learn all necessary transformations directly from the data and which can perform natural language generation with some amount of variety in phrasing. POV conversion bears similarities to the machine translation \cite{edunov2018understanding}, paraphrasing \cite{witteveen2019paraphrasing}, text generation \cite{gatt2018survey}, abstractive summarization \cite{GUPTA201949}, and question-and-answer \cite{zhang2020retrospective} tasks, all of which have performed well using architectures that have either an encoder, a decoder, or both.

As a basic benchmark for encoder-decoder sequence to sequence (Seq2Seq) model, we first consider a classic ``LSTM-LSTM" model with dot product attention \cite{sutskever2014sequence, bahdanau2014neural}. From there, we tried a variety of encoders, decoders, and attention. A quick modification from this classic structure is a transformer block as the encoder \cite{vaswani2017attention}. A decoder structure that seems particularly well suited for our task was CopyNet, which recognizes certain tokens to be copied directly from the input and injected into the output \cite{gu2016incorporating}.

As of the time of writing, high-performing systems for many NLP tasks are based on transformer architectures \cite{vaswani2017attention,devlin2018bert,radford2019language,lample2019crosslingual} that are first pretrained on large corpora of unlabeled data using masked language modeling and its variants, next sentence prediction and its variants, and related tasks. The models are then fine-tuned on the desired downstream task. The Text To Text Transfer Transformer (T5) model \cite{raffel2019exploring} is our choice of encoder-decoder transformer, which achieved state of the art performance on SQuAD, SuperGLUE, and other benchmarks in 2019.

We are not aware of prior work specifically targeted at messaging point of view conversion for virtual assistants. This initial investigation into perspective switching begins to formulate what people frequently do in natural conversations. By extending this formulation to VAs, we provide mechanisms to parse out messy natural speech and maximize informational gain. Identifying perspectives associated with a segment of natural speech may help perspective unification for pre-processing and anaphora resolution. Moreover, this could benefit additional tasks such as quotation detection \cite{papay-pado-2019-quotation}, contextual paraphrase generation, and query re-writing \cite{lin2020query,yu2020fewshot}.


\section{Problem Statement}\label{prob}

Given our assumption that the user name, contact name, and message content are known, our objective is to convert the POV of the voice message.
Whether each step is performed explicitly, as with the rule-based model, or whether the model learns them, as with Seq2Seq models, the POV conversion in our design subsumes the following sub-tasks.

When the VA relays a message from Joe to Jill, the source contact name, Joe, is a crucial piece of information for Jill. 
Yet it is often missing from the input. 
The first step of our POV conversion model, therefore, is to add the name of the source contact to the output utterance. 
On the other hand, the contact name, Jill, is a redundant piece of information for Jill herself and can optionally be removed.
Note that with our dataset we do include the contact name in the converted output, because we assume that the VA is a communal device with multiple users (thereby making the contact name relevant even in the converted output).

Voice messages, like all sentences, come in different varieties and perform different functions. 
They can be used to give a statement, make a request, or ask a question \citep{austin1975things}. 
In our rule-based model, we conducted a classification task to categorize messages into different types. 
Accordingly, we needed to pair appropriate reporting verbs, i.e., verbs used to introduce quoted or paraphrased utterances, with distinct message types. 
Messages used to inquire (\textit{are you coming for dinner}) work better with the verb \textit{ask} (\textit{Joe asks if you are coming for dinner} or \textit{Joe is asking if you are coming for dinner}), whereas messages used to make an announcement (\textit{dinner's ready}) work better with the verb \textit{say} (\textit{Joe says dinner's ready}).

A POV change often leads to changes in pronouns. 
This is illustrated in Table \ref{PronominalChange}. 
Among other things, our model needs to be able to convert a first person (\textit{I}) to a third person (\textit{he}), and a third person (\textit{he}) to a second person (\textit{you}). 
Accompanied with the change of pronoun is a change in verb forms (i.e., \textit{am} to \textit{is}, and \textit{is} to \textit{are}), since the grammar requires that the verb and its subject agree in person.
\begin{table}[h!]
\centering
\caption{Examples of Pronominal Change}
\label{PronominalChange}
\begin{tabular}{ll}
\hline
\multicolumn{2}{c}{Example 1} \\
\hline 
Input & (Tell Bob) \textbf{I} am running late \\ 
Output & (Joe says) \textbf{He} is running late  \\ 
\hline
\multicolumn{2}{c}{Example 2} \\
\hline 
Input & (Ask Bob) if \textbf{he} is coming for dinner \\ 
Output & (Joe asks) if \textbf{you} are coming dinner \\ 
\hline 
\end{tabular} 
\end{table}

The final step is to reverse subject-auxiliary inversion. 
When the voice message is a direct question, such as \textit{ask Bob are you coming for dinner}, the ideal output would be \textit{Joe asks if you are coming for dinner}. 
In this case, our model needs to be able to convert a direct question to an indirect one. 
Such a transformation involves reversing the relative order between the main verb (or auxiliary) and its subject.
In our rule-based model we used a Part Of Speech (POS) tagger to distinguish direct questions from indirect questions, and a constituency parser to identify the subject and the main verb of each message.

\section{Data}\label{data}

Data collection and verification was performed both by Amazon Mechanical Turk workers \cite{callison2010creating} and by internal associates who were not part of our research group and had no knowledge of the system.
We used separate campaigns to achieve statistical significance across the major categories of utterances (See Table \ref{MessageTypes}).
In all cases, workers were asked first to create reasonable and realistic utterances (as if spoken to the VA).
They were then asked to convert those utterances into a natural and faithful output from the VA.
The data were post-processed using a few assumptions:
\begin{itemize}
    \item{The names were replaced with special tokens, being \verb|@CN@| for contact name and \verb|@SCN@| for source contact name.}
    \item{When there were ambiguous pronouns in the input sentence in the third person, the pronouns were assumed to be referencing the contact, and not an outsider 3rd person.}
    \item{When there were gender ambiguities in the singular 2nd person pronoun in the input, the conversions used the gender neutral ``they." Other researchers have devised methods to use the correct pronoun based on lookup tables \cite{malmi2019encode}, but such was beyond the scope of our project.}
\end{itemize}

Our total dataset is composed of 46,565 samples, and we used a 70/15/15 split for training/validation/test. We have released our dataset publicly \footnote{https://github.com/alexa/alexa-point-of-view-dataset}.

\section{Evaluation Metrics}

We sought to evaluate both the \textit{faithfulness} and the \textit{naturalness} of the outputs from our models. Faithfulness is the degree to which the model's output correctly preserves both the meaning of the message and the fact that the voice assistant is conveying a message from another user to the recipient. Naturalness is the degree to which the final output sounds like something that a real person might say. 

To automatically evaluate faithfulness, we considered three ``off the shelf'' evaluation algorithms. BiLingual Evaluation Understudy (BLEU) is an evaluation algorithm focused on the precision of the model's output \cite{papineni2002bleu}. Recall-Oriented Understudy for Gisting Evaluation (ROUGE) was later developed with a focus toward recall \cite{rouge}, though for our work, we considered the ROUGE-L F1 score, which considers the longest substrings of matches between model outputs and ground truth, and which balances precision and recall (by using the F1 score). METEOR addresses some of the shortcomings of BLEU and ROUGE by allowing for matches of stemmed words, synonyms, and paraphrases \cite{denkowski:lavie:meteor-wmt:2014}. With our data, synonyms and paraphrases are quite common in the carrier phrase (\textit{Bob would like to inform you that} vs \textit{Bob would like to tell you that}). Ultimately we chose to use BLEU, given its popularity and familiarity across the field, as well as METEOR, given its ability to handle stemming, synonyms, and paraphrasing. BLEU is calculated using the \texttt{corpus\_bleu} method from the \texttt{nltk.translate} package, and METEOR is calculated by averaging the \texttt{single\_meteor\_score} values, also from the \texttt{nltk.translate} package \cite{BirdKleinLoper09}.

As another gauge of faithfulness, we considered the cosine similarity between the sentence embeddings of a given model's output and the corresponding ground truth. We used pretrained fastText embeddings \cite{bojanowski2016enriching}, as well as the \texttt{spatial} module from the \texttt{SciPy} library \cite{2020SciPy-NMeth}. Sentence-level cosine similarity was least correlated with BLEU, as seen in Table \ref{correlations}, but its variance was quite small, with values ranging 0.93 to 0.96 for T5. Likely, the metric has been pretrained on a large corpus of many, generalized domains, and therefore, does not adequately capture messaging-specific variance.

\begin{table}[h!]
\centering
\caption{The correlation between the relative changes in the \textit{faithfulness} metrics as taken from the T5 validation curves.}
\label{correlations}
\resizebox{\linewidth}{!}{%
\begin{tabular}{cccc}
\hline 
&\textbf{ROUGE-L F1}&\textbf{METEOR}&\textbf{Cosine Sim}\\ 
\hline 
BLEU&0.87&0.67&0.63\\
ROUGE-L F1&&0.91&0.77\\
METEOR&&&0.78\\ 
\hline
\end{tabular} 
}
\end{table}

For naturalness, we first downloaded the GPT model \cite{radford2018improving} using the \texttt{transformers} library \cite{Wolf2019HuggingFacesTS}. We ran each sample through the GPT model and calculated word-count-normalized perplexity based on exponentiation of the model's loss, which is already normalized according to post-tokenization token count. In all cases, we substituted \texttt{bob} for \texttt{@cn@} and \texttt{john} for \texttt{@scn@}. For each sample, we calculated the relative perplexity be dividing the ground truth perplexity by the model's perplexity. Since a lower perplexity is better, this means that a higher relative perplexity corresponds to better performance by the model. To calculate corpus-level relative perplexity, we simply calculated the mean of the relative perplexity for each sample.

Finally, a small subset of model output was human evaluated for faithfulness as a binary metric and naturalness as a semi-binary metric. Since naturalness is highly subjective, dependent on regionality or grammaticality of the speaker, it was evaluated on a 1 to 4 scale, with 1 being unacceptable and 4 being perfectly fluent. Then, this scale was converted to a binary metric. The full human evaluation is detailed in the released dataset.

\section{Rule-Based Model}

\begin{figure}[h!]
\resizebox{\linewidth}{!}{%
 \includegraphics[scale=1]{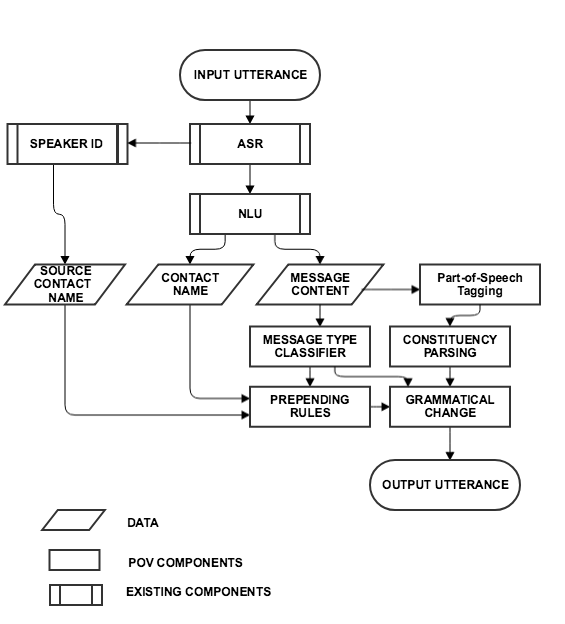}
}
  \caption{The end-to-end architecture of the rule-based model}\label{workflow}
\end{figure}

We represent the overall design of the rule-based system in Figure \ref{workflow}. 
Our model ingests raw voice messages as input.
The messages are first transcribed by ASR, and then go through NLU for intent classification and slot detection.
At this point of our design, three slots would be determined by existing components, namely, Source Contact Name (the sender), Contact Name (the recipient) and Message Content.
We submit these slots to the POV component, to be discussed in detail in the following subsections.
The final product of our model is an utterance that switches its POV from the sender of the message to the VA.

\subsection{Message Classification}\label{classification}
We start off by classifying the messages into different types.
Inspired by \cite{briggs2017enabling}, we define the different message types as in Table \ref{MessageTypes}.
The variables $\alpha$ and $\beta$ denote the speaker and the addressee, respectively.
\begin{table}[h!]
\centering
\caption{Four Types of Voice Messages}
\label{MessageTypes}
\resizebox{\linewidth}{!}{%
\begin{tabular}{ll}
\hline 
\textbf{Name} & \textbf{Explanation}  \\ 
\hline 
\multirow{3}{*}{Stmt($\alpha$,$\beta$,X)} & denotes a statement by $\alpha$ to $\beta$ \\ 
 & asserting X is true \\
 & E.g., \textit{Tell Bob dinner is ready}\\
\hline 
\multirow{3}{*}{AskYN($\alpha$,$\beta$,X)} & denotes a question by $\alpha$ to $\beta$ \\
 & inquiring if X is true \\ 
 & E.g., \textit{Ask Bob if dinner is ready}\\
\hline 
\multirow{4}{*}{AskWH($\alpha$,$\beta$,X)} & denotes a question by $\alpha$ asking $\beta$\\
 &  to resolve the reference specified \\
  & by X (e.g., location, identity, etc.)\\ 
  & E.g., \textit{Ask Bob when dinner will be ready}\\
\hline 
\multirow{3}{*}{Req($\alpha$,$\beta$,X)} & denotes a request by $\alpha$ to $\beta$  \\
 & to perform action X\\ 
 & E.g., \textit{Ask Bob to join us for dinner}\\
\hline 
\end{tabular} 
}
\end{table}

\begin{figure}[h!]
\resizebox{\linewidth}{!}{%
\includegraphics[scale=1]{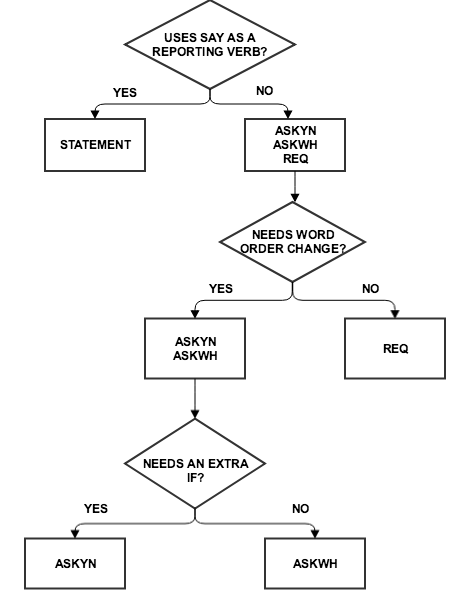}
}
\caption{Four Message Types Correspond to Four Types of Conversion Rules}\label{decisiontree}
\end{figure}

We perform classification because each message type requires slightly different follow-up procedures for POV conversion.
The reporting verb \textit{say} is only used with Stmt messages (\textit{Joe says dinner's ready}) whereas \textit{ask} is compatible with the rest.
AskYN and AskWH messages may involve changing a direct question to an indirect one (\textit{Joe asks when dinner's ready}), whereas the other two do not.
In addition, an extra \textit{if} needs to be inserted for AskYN messages after the grammatical change (\textit{Joe asks if dinner's ready}).
This is summarized by the decision tree in Figure \ref{decisiontree}.

Given our definition for the four message types, we have the following observations.
\begin{itemize}
\item AskWH messages include wh-words such as \textit{who, what, when, where}, etc. 
\item AskYN messages include phrases like \textit{ask if}, \textit{ask whether (or not)}, or questions that start with auxiliaries like \textit{are, is, can, will}, etc.
\item Req messages include phrases like \textit{tell to}, \textit{ask to}, \textit{remind to}, etc.
\item Stmt messages are a mixed bag. Phrases that belong to this broad group include, but are not limited to, \textit{tell that}, \textit{message that}, \textit{remind that}, etc.
\end{itemize}
In order for these phrases to be included as model features, we use n-grams ranging from 1 to 5 tokens in length.

We considered a training set 5,992 samples and a validation set of 927 samples, each of which was a subset of the main dataset for which the message category had been human-annotated. To reduce the weights of the common words that occur across message types, we used Term Frequency-Inverse Document Frequency (TF-IDF) \citep{park2012literature,leskovec2014mining}.
Many of the traditionally-used English stop words are actually crucial phrases that are necessary for our classification, so we created our own list of stop words that includes common first names, prepositions, articles (\textit{the, a, an}), and filler words (\textit{please, um}).
Between TF-IDF-based thresholding and our stop word list, we reduced the number of classification features to 188.
We then trained a linear Stochastic Gradient Descent (SGD) classifier with modified Huber loss, L2 regularization, and 5,000 iterations. \citep{robbins1985stochastic}.

\subsection{POS Tagging and Constituency Parsing}\label{constituency}
Voice messages are also tagged and parsed using the Stanford CoreNLP package \citep{klein2003accurate,klein2003fast,zhu2013fast}.
Constituency parsing allows for sentences to be syntactically analyzed, and it parses sentences into subphrases with words as terminal nodes.
Dependencies and tree structures are parsed via a pre-trained neural network that takes words and POS tagged inputs and outputs relations between nodes.
The POS information is used as labels for each node in the constituency tree.

The POS tagger \citep{marcus1993building} helps us distinguish direct questions from indirect ones.
The distinction between the two types of questions and their respective constituents is illustrated in Figure \ref{constituency}, and an index of the POS tags used are shown in Table \ref{postag} \citep{bies1995bracketing}.
Only direct questions carry the POS label SQ (for inverted questions).
The constituency parser identifies the subject and the main auxiliary of each utterance.
They are the first two daughters of the embedded S or SQ.
\begin{figure}[h!]
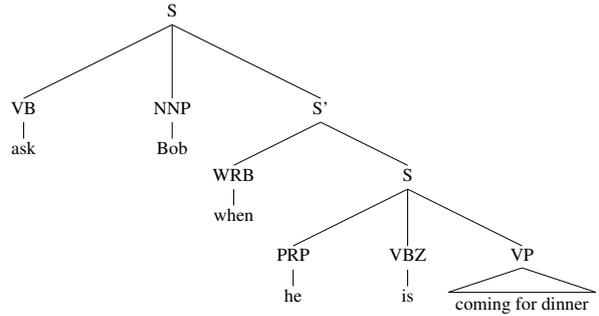

a. AskWH + direct question\\
\resizebox{\linewidth}{!}{%
\Tree [.S [.VB ask ] [.NNP Bob ] [.S' [.WRB when ] [.SQ [.VBP are ] [.PRP you ]  \qroof{coming for dinner}.VP ] ] ]
}
b. AskWH + indirect question\\
\resizebox{\linewidth}{!}{%
\Tree [.S [.VB ask ] [.NNP Bob ] [.S' [.WRB when ] [.S [.PRP he ] [.VBZ is ]  \qroof{coming for dinner}.VP ] ] ]
}
\caption{Constituency trees of direct vs. indirect question}\label{constituency}
\end{figure}
\begin{table}[h!]
\centering
\caption{Index of the Part Of Speech (POS) Tags}\label{postag}
\resizebox{\linewidth}{!}{%
\begin{tabular}{ll}
\hline 
\textbf{POS Tag} & \textbf{Information }\\ 
\hline 
S & simple declarative clause \\ 
SBAR (or S') & clause introduced by a subordinating conjunction \\ 
SQ & Inverted yes/no question, or main clause wh-question \\ 
NNP & proper noun, singular \\
VB & verb, base form \\ 
VBP & verb, non-3rd person singular present\\ 
VBZ & verb, third person singular present \\ 
WRB & wh-adverb\\
PRP & personal pronoun\\
VP & verb phrase\\
\hline
\end{tabular} 
}
\end{table}

The sentence level tags (S, S', SQ) are crucial in determining whether the word order between the subject and the auxiliary is reversed.
The word level tags (VB, VBP, VBZ), on the other hand, indicate which verb form needs to be changed as part of the POV conversion.


\subsection{Transformations}\label{deterministic}
After POS tagging and constituency parsing are complete, our rule service proceeds as follows.
\begin{enumerate}
\item  \textbf{Searching for missing contact name in message content}. In most cases the contact name is given by the existing NER component, but occasionally it is missing from the input. For instance, if Joe says, \textit{Find out if Nate is bringing anything to the party}, NER would label \textit{Nate is bringing anything to the party} as the message content. The contact name, Nate, is hidden inside the message content. Since NER cannot provide any given slot with two labels, in cases like this, we employ rules to recover the embedded contact.
\item \textbf{Changing word order}. This step only applies to direct questions in AskYN and AskWH messages. During this process, multiple types of grammatical changes may apply, including do-deletion, and subject-auxiliary reversal (\textit{are you} $\rightarrow$ \textit{you are}).
\item \textbf{Swapping pronouns/contact names}. We use rules to convert a first person (\textit{I}) to a third person (\textit{he/she}) and a third person (\textit{he/she}) to a second person (\textit{you}). In cases where the contact name resides inside the message content, the rules would find it and switch it with a second person pronoun.
\item \textbf{Fixing verb agreement}. This step is to make sure the main verb/auxiliary agrees with the converted subject pronoun in person and number. In sentences with present tense, if we switch the subject \textit{she} to \textit{you}, we must change the main verb to its base form (\textit{is} $\rightarrow$ \textit{are}, \textit{wants} $\rightarrow$ \textit{want}, VBZ $\rightarrow$ VBP) as well.
\item \textbf{Adding prepending rules to reconstructed message content}. Finally we add the source contact name and appropriate reporting verbs to the beginning of each output, among other things. Each type of message has a different set of prepend rules and the VA can randomly choose prepends in the same set to sound more spontaneous. For example, an AskYN message with a direct question would need a prepend rule like\textit{@SCN@ asks if} or \textit{@SCN@ is wondering whether}. Similarly, a Req messages might use \textit{@SCN@ asks you to} or \textit{@SCN@ would like to remind you to} as prepends.
\end{enumerate}
Table \ref{POVChange} illustrates how our model deals with various types of voice messages.

\begin{table}[h!]
\centering
\caption{Examples of POV Conversion Results. ``Teresa'' is the name of the Source Contact Name (the sender).}
\label{POVChange}
\resizebox{\linewidth}{!}{%
\begin{tabular}{ll}
\multicolumn{2}{c}{Example 1: Stmt} \\
Input & Can you let mom know that I finally mailed her package? \\  
Output & Teresa says she finally mailed your package.  \\ 
\multicolumn{2}{c}{Example 2: AskYN + direct question} \\ 
Input & Ask Haley can I borrow your juicer?  \\  
Output & Teresa asks if she can borrow your juicer \\  
\multicolumn{2}{c}{Example 3: AskYN + indirect question} \\ 
Input & Can you ask Blade if he's still having a party tomorrow \\  
Output & Teresa asks you if you're still having a party tomorrow\\  
\multicolumn{2}{c}{Example 4: AskWH + direct question} \\ 
Input & Text alyssa what type of wine do you want \\  
Output & Teresa asks what type of wine you want \\  
\multicolumn{2}{c}{Example 5: AskWH + indirect question} \\ 
Input & Ask Jeff what he's doing tonight \\  
Output & Teresa asks what you are doing tonight \\  
\multicolumn{2}{c}{Example 6: Req} \\ 
Input & Text Will to grab some apples on his way home \\  
Output & Teresa asks you to grab some apples on your way home \\  
\multicolumn{2}{c}{Example 7: Missing contact} \\ 
Input & Find out if Nate is bringing anything to the party \\  
Output & Teresa asks if you are bringing anything to the party \\  
\end{tabular} 
}
\end{table}

\section{LSTM-LSTM Model}

The first Seq2Seq variant, which we call ``LSTM-LSTM,'' was implemented using  the AllenNLP library \cite{Gardner2017AllenNLP}.   Byte Pair Encoding (BPE) was used for tokenization to reduce out-of-vocabulary errors \cite{DBLP:journals/corr/SennrichHB15}.  For instance, the word ``\verb|abdicated|" is processed into multiple tokens of subwords, ``\verb|ab dic ated|."  The wikipedia pre-trained \texttt{BPEmb} package was used \cite{heinzerling2018bpemb}, and the minimum frequency for the vocabulary was set to 3.  The model consists of 256-dimensional word embeddings, a Long Short Term Memory (LSTM) encoder, dot product attention, a hidden representation of 256 dimensions, and an LSTM decoder. We used ADAM \cite{kingma2014adam} as our optimizer and a beam size of 8 for decoding. Work was performed on an AWS ml.p2.8xlarge instance, which includes 8 NVIDIA Tesla K80 GPUs.

\section{Transformer-LSTM Model}

As our second variant, we again used the AllenNLP library, but for our encoder, we used a single transformer block composed of 8-headed self-attention and a 128-dimension fully-connected network. The decoder was the same LSTM as with the LSTM-LSTM model, with the same beam size of 8. We again used ADAM as the optimizer, as well as an AWS ml.p2.8xlarge instance.  The data were tokenized using BPE as shown above.

\section{CopyNet Model}

The main disadvantage of the neural machine translation approaches, compared to the rule based approach, is the distortion of the message payload, including insertions and deletions of content-significant but rare words. The faithfulness of the message deteriorates, though the construction of the messages becomes more flexible.  To mitigate for this, the ideal architecture should implicitly distinguish which tokens are messages, which tokens indicate the semantic structure classification, and where in the sentence those tokens are, while still leveraging the fluency and naturalness of neural language generation.  One such model is CopyNet.

CopyNet identifies input token arrangements to copy or modify according to a ``Copy and Generate'' strategy in its decoder.  The LSTM encoder output is first fed through an attentive read layer, which encodes the source into a sequence of short-term memory.  This short-term memory is assessed using ``Copy mode'' and ``Generate mode,'' which identify tokens that should be reproduced as-is in the output and tokens that should be generated.  Then, a prediction $s_t$ is composed using this hybrid probability.  In addition to the attentive read, CopyNet also selectively reads location and content addressed information from the previous predictions, $s_{t-1}$, along with word embeddings.  In combination, this model learns from the encoded source what to copy, what to generate, and where, and it strategically composes the message in its decoder.

AllenNLP's implementation of CopyNet was used.  Similar to other LSTM based approaches, the input data were tokenized using BPE.  A single layer LSTM encoder with 128 hidden dimensions and dot product attention were specified, with SGD as the optimizer.  The model training was done on an AWS ml.p2.8xlarge instance, which includes 8 NVIDIA Tesla K80 GPUs.

\section{Transformer-CopyNet Model}

Transformer encoder was investigated further to boost performances on CopyNet Model.  An 8-layer stacked self-attention layer was used as an encoder in addition to AllenNLP's implementation of CopyNet, with Adam optimizer. Instead of using BPE for preprocessing the input, the raw input tokens were used to feed into the transformer encoder directly. Using the transformer encoder also had an additional benefit of cutting the training epoch by more than half. For 6-10 layer stacked self-attention encoder and CopyNet decoder, the model typically achieved optimum output around 20-23 epochs.
%
%

\section{T5 Model}

We started with the T5 base model, which had been pretrained on the Colossal Clean Crawled Corpus (C4), a  dataset derived from website content, using the text infilling task. The model is composed of 220 million parameters across 12 blocks, where each block is composed of self-attention, optional encoder-decoder attention, and a feedforward network. Pretraining occurred over 524k steps, where each step is a batch of 128 samples with a maximum sequence length of 512.

To fine tune the model, we formatted our samples into the continuous text format that the model expects, i.e. \texttt{
b'pov input: please invite @cn@ to come over tonight'
} and 
\texttt{
b'hi @cn@, @scn@ wants you to come over tonight'
}.

Although the T5 authors fine-tuned for 262k steps using a batch size of 128 samples, we found that 60k global steps (63 epochs) was sufficient for our task. In fact, performance degraded above 76k global steps. See Figure \ref{val_curves}. We used a constant learning rate of 0.003 during fine-tuning. Dropout was set to 0.1. Some work was performed on a 72-core AWS EC2 instance using Intel Xeon Platinum 8000 series processors with 144 GB of RAM (ml.c5d.18xlarge) and some on an AWS ml.p3.16xlarge instance, which includes 8 NVIDIA Tesla V100 GPUs.

\section{Results}\label{results}

\subsection{Rule-Based Model Classification Results}

The classification results for the message types are shown in Table \ref{LinearModelPerformance}. Performance is generally good across all message types, with F1 scores ranging from 0.91 to 0.98.

\begin{table}[h!]
\centering
\caption{The performance of the message type linear SGD classifier used in the rule-based model.}
\label{LinearModelPerformance}
\begin{tabular}{lccc}
\hline 
&\textbf{Precision}&\textbf{Recall}&\textbf{F1}\\ 
\hline 
Stmt&0.95&0.94&0.94\\
AskWH&0.97&0.99&0.98\\
Req&0.91&0.92&0.91\\
AskYN&0.96&0.94&0.95\\ 
\hline
\end{tabular} 
\end{table}

\subsection{Validation Summary}

\begin{figure*}[ht]
\centering
\begin{subfigure}{0.32\textwidth}
\centering
\includegraphics[width=\textwidth]{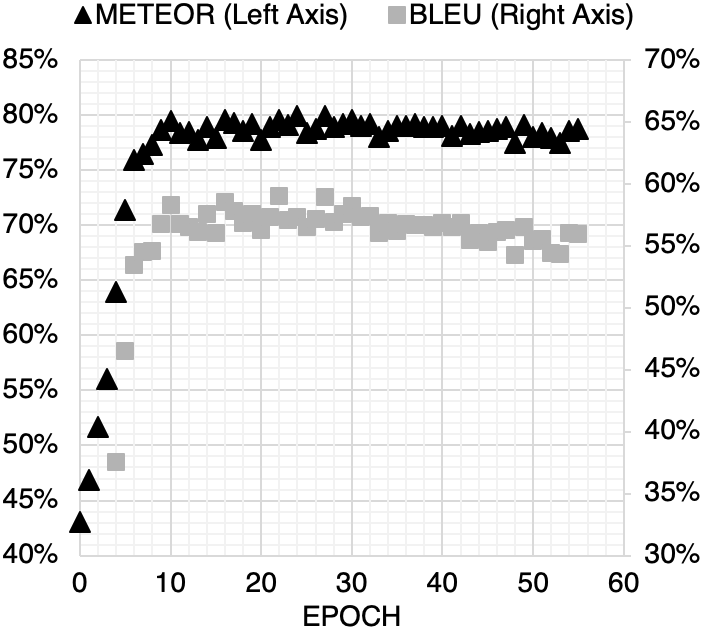}
\caption{LSTM-LSTM}
\end{subfigure}
\begin{subfigure}{0.32\textwidth}
\centering
\includegraphics[width=\textwidth]{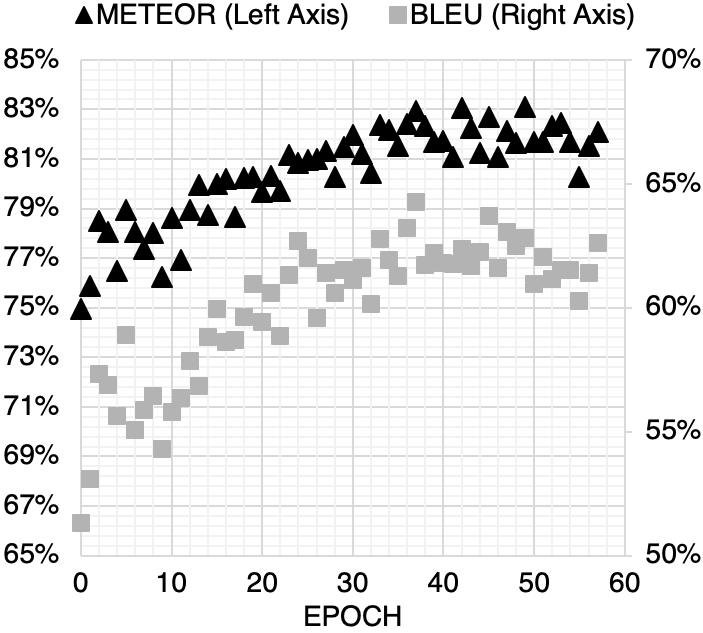}
\caption{CopyNet}
\end{subfigure}
\begin{subfigure}{0.32\textwidth}
\centering
\includegraphics[width=\textwidth]{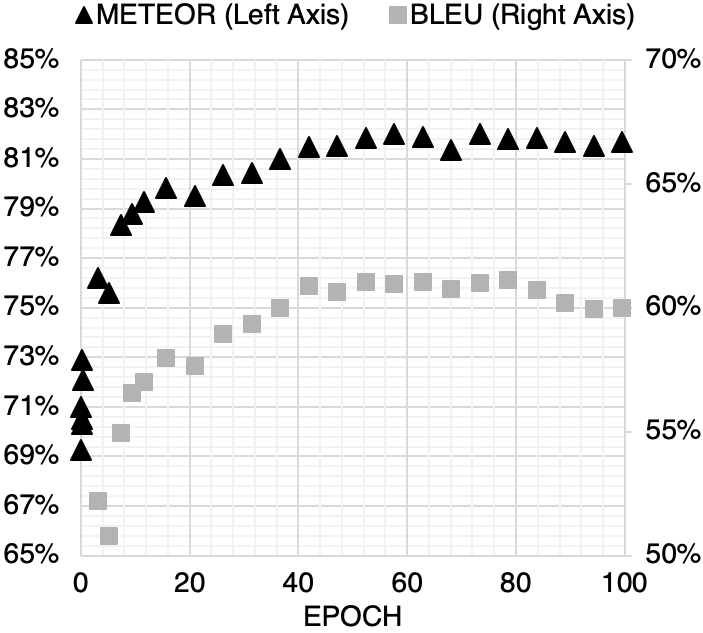}
\caption{T5}
\end{subfigure}
\caption{BLUE and METEOR validation data for the LSTM-LSTM, CopyNet, and T5 models. Greater model complexity requires more training time.}
\label{val_curves}
\end{figure*}

\begin{figure*}[ht]
\centering
\begin{subfigure}{0.32\textwidth}
\centering
\includegraphics[width=\textwidth]{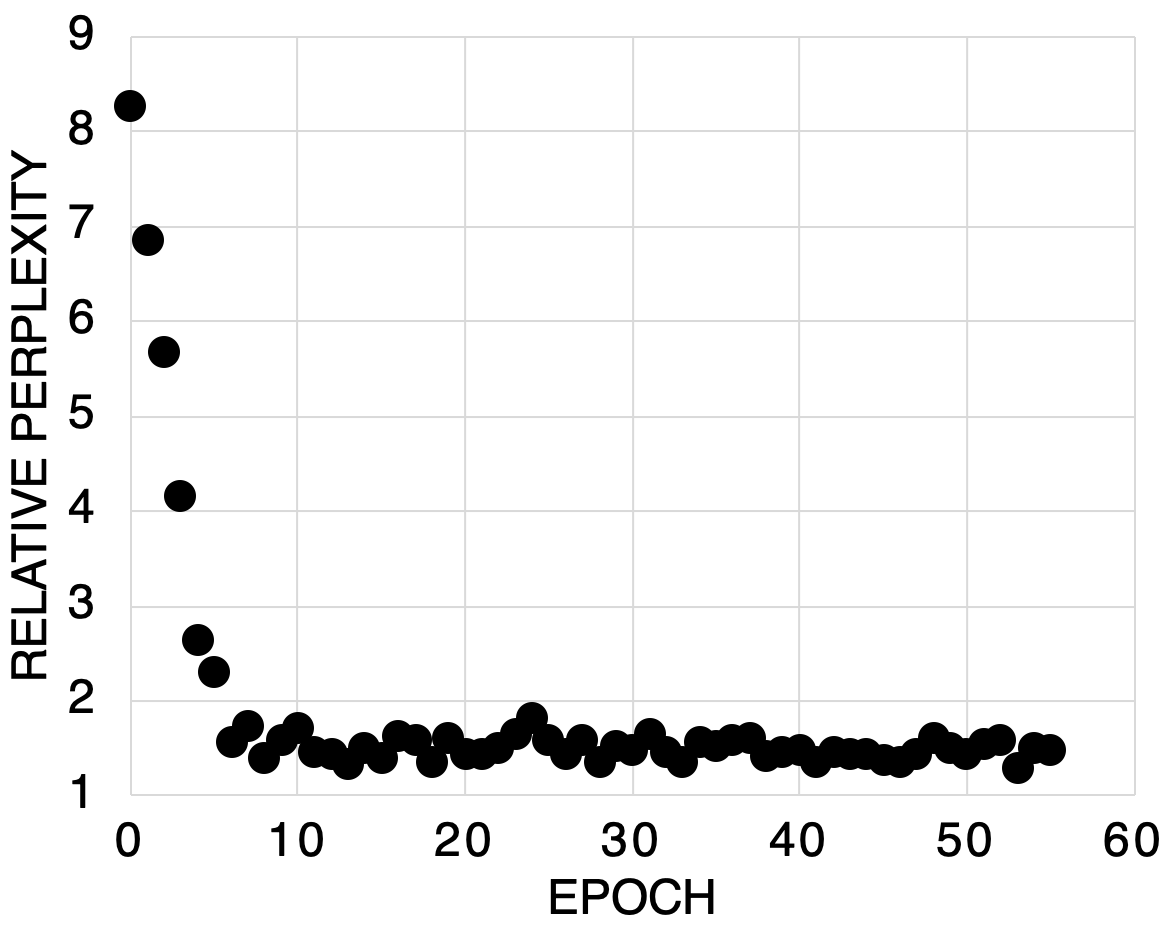}
\caption{LSTM-LSTM}
\end{subfigure}
\begin{subfigure}{0.32\textwidth}
\centering
\includegraphics[width=\textwidth]{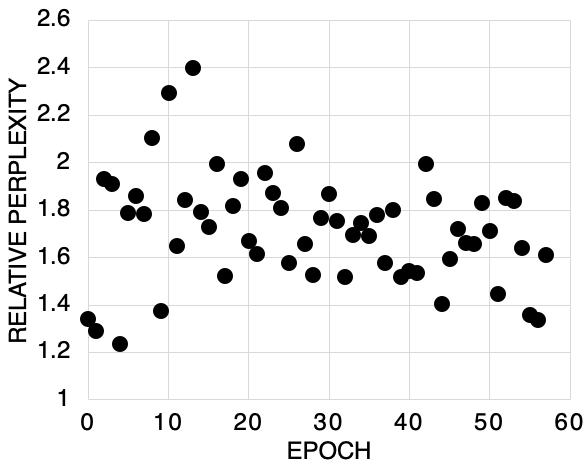}
\caption{CopyNet}
\end{subfigure}
\begin{subfigure}{0.32\textwidth}
\centering
\includegraphics[width=\textwidth]{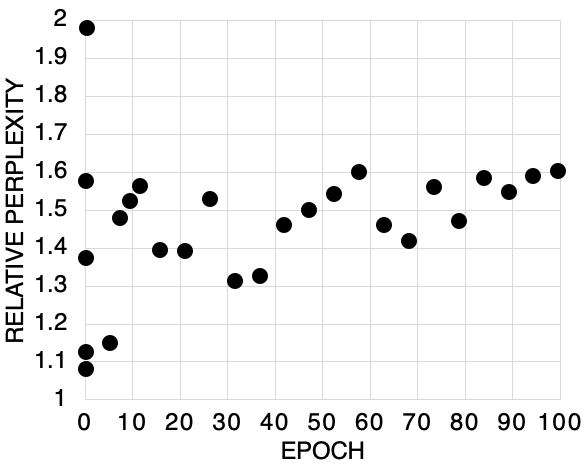}
\caption{T5}
\end{subfigure}
\caption{The relative perplexity validation data from the LSTM-LSTM, CopyNet, and T5 models. A higher relative perplexity is better. A decaying exponential trend is seen in the LSTM-LSTM model's data. Though CopyNet and T5 data are noisy, a downward trend is evident the CopyNet results, and convergence is evident in the T5 results. All data are above 1, which means that the model output was more natural (with lower absolute perplexity via the GPT model) than the ground truth samples on average.}
\label{relative_perplexity_curve}
\end{figure*}

Faithfulness validation curves are shown in Figure \ref{val_curves}, and validation results of relative perplexity in Figure \ref{relative_perplexity_curve}.

A clear decaying exponential relationship between relative perplexity and training epoch can be seen with the LSTM-LSTM model. For the CopyNet and T5 models, the relative perplexity data were quite noisy, likely because the model's learning rate and regularization are tuned for the loss function and for the faithfulness validation metrics. Convergence is especially discernable as training progresses for the T5 model.

Initially surprisingly, relative perplexity is always greater than 1 across seq2seq models' results, which means that the model's output is more natural than the ground truth data on average, at least according to the GPT model used to evaluate perplexity. Moreover, relative perplexity generally worsens as training progresses for the LSTM-LSTM and the CopyNet models. The reason for this behavior is qualitatively explainable by examining the data. For example, in the first epoch of training with the LSTM-LSTM model, the model hypothesized \textit{hi bob john would like to know if you are going}, whereas the ground truth was \textit{hi bob john is asking if you are from germany}. The model has not yet learned to be faithful, and it is instead hypothesizing high probability outputs which are of low absolute perplexity. Indeed, this behavior continues for all Seq2Seq models even after training is complete. The models choose carrier phrases that are most likely and most natural, such as \textit{john is asking}, as opposed to some of the more varied, lower probability ground truth carrier phrases like \textit{john is requesting to know}.

\subsection{Overall Results}

Results on the held out test set for all of our models are given in Table \ref{vanilla}. In order to prevent any data leakage, evaluation on the final test set was only conducted once per model as the last step of our study. T5 performed the best for BLEU (63.8) and METEOR (83.0), whereas CopyNet had the best relative perplexity (1.59). A small scale human evaluation was complete on T5 and CopyNet as well, and it seems to corroborate our choice of automatic metric for faithfulness. Both performed similarly on naturalness, with accuracy of 97\% for CopyNet and 98\% for T5. T5 outperformed on faithfulness at 98\%, whereas CopyNet achieved 94\%. The human evaluation result for naturalness is a bit subjective, but T5 model's slight edge on faithfulness reflects its precision of relayed message content.

\begin{table}[h!]
\centering
\caption{Results for all models using a held-out test set of 6,986 samples, including the quantity of parameters in the model (Params), the corpus BLEU score, the average METEOR score, and the relative perplexity (higher is better).}
\label{vanilla}
\resizebox{\linewidth}{!}{%
\begin{tabular}{lcccc}
\hline 
\textbf{Model}&\textbf{Params}&\textbf{BLEU}&\textbf{METEOR}&\textbf{Perpl}\\ 
\hline 
Rule-based&-&46.6&72.3&0.918\\
LSTM-LSTM&5.3M&55.7&78.1&1.39\\ 
Tsfmr-LSTM&\textbf{4.9M}&50.9&75.0&1.39\\
CopyNet&5.3M&63.1&82.0&1.54\\
Tsfmr-CopyNet&5.9M&63.7&82.8&\textbf{1.59}\\
T5 &220M&\textbf{63.8}&\textbf{83.0}&1.42\\
\hline
\end{tabular} 
}
\end{table}

\section{Conclusion}\label{conclusion}
In this paper, we considered the task of converting the point of view of an utterance. We designed a system that takes in raw voice messages and generates natural and faithful outputs with a changed point of view.
To the best of our knowledge, this is the first effort to convert the point of view of messages directed to virtual assistants.
T5 and Transformer-CopyNet performed similarly on BLEU and METEOR. 
T5 had a slight edge on BLEU by 0.16\% and METEOR by 0.24\%.
On the other hand, Transformer-CopyNet had a significantly better relative perplexity, by 12.0\%, indicating that the model is much more fluent.
Given that T5 is 37 times larger than CopyNet and Transformer-CopyNet seems to reach optimal performance in 20-23 epochs, we recommend CopyNet for similar tasks.
Since CopyNet uses vocabulary constructed entirely from the training set, the faithfulness based metrics could be improved upon by leveraging pretrained word embeddings for reducing OOV errors.
T5 seems to have a slight edge on faithfulness-based metrics, which could imply a larger pre-trained T5 may achieve a higher level of message content lexical similarity.
In that case, there are even larger pretrained T5 models that could be leveraged, with the largest containing 11 billion parameters.

We are optimistic that future experimentation will yield even better results. CopyNet's decoding strategy should be investigated further, as well as different types of encoding strategies, such as a pretrained word embeddings like BERT or Semantic Role biased encodings \cite{DBLP:journals/corr/abs-1804-08313}, which may provide the decoder with more linguistically salient information.  Moreover, these word embeddings could be pre-tuned for our tasks specifically.  Addititionally, the feedforward decoder variant of the \texttt{\textsc{LaserTagger}} model \cite{malmi2019encode} shows promising results on similar tasks while maintaining a single-sample inference latency of 13 ms (using a NVIDIA Tesla P100).

Ultimately, we envision a future in which virtual assistants can serve as human-like communication intermediaries between two users and even groups of users, particularly over long messaging chains where context would be paramount.
Point of view conversion is a small step toward that vision.



\bibliography{pov}
\bibliographystyle{acl_natbib}

\end{document}